\documentclass[10pt,twocolumn,letterpaper]{article}

\usepackage[pagenumbers]{cvpr} 

\usepackage{bm}
\usepackage{microtype}
\usepackage{multirow}
\usepackage{graphicx}
\usepackage{xcolor}
\usepackage{color}
\usepackage[accsupp]{axessibility}  

\newcommand{\tbf}[1]{\textbf{#1}}
\newcommand{\tul}[1]{\underline{#1}}
\newcommand{\tworow}[2]{\begin{tabular}[c]{@{}c@{}}#1\vspace{-2pt}\\#2\end{tabular}}

\newcommand{\psp}{\kern0.2ex}
\newcommand{\nsp}{\kern-0.1ex}

\DeclareMathOperator*{\argmin}{arg\,min}

\definecolor{cvprblue}{rgb}{0.21,0.49,0.74}
\usepackage[pagebackref,breaklinks,colorlinks,allcolors=cvprblue]{hyperref}

\def\paperID{2025} 
\def\confName{CVPR}
\def\confYear{2025}

\title{Continuous Locomotive Crowd Behavior Generation}

\author{Inhwan Bae, Junoh Lee and Hae-Gon Jeon\footnotemark[1]\\
Gwangju Institute of Science and Technology, South Korea\\
{\tt\small \{inhwanbae, juno\}@gm.gist.ac.kr, haegonj@gist.ac.kr}}

\begin{document}

\renewcommand*{\thefootnote}{\fnsymbol{footnote}}
\twocolumn[{%
\renewcommand\twocolumn[1][]{#1}%
\maketitle
\vspace{-10mm}
\begin{center}
    \centering
    \captionsetup{type=figure}
    \includegraphics[width=\linewidth,trim={0 7mm 0 7mm},clip]{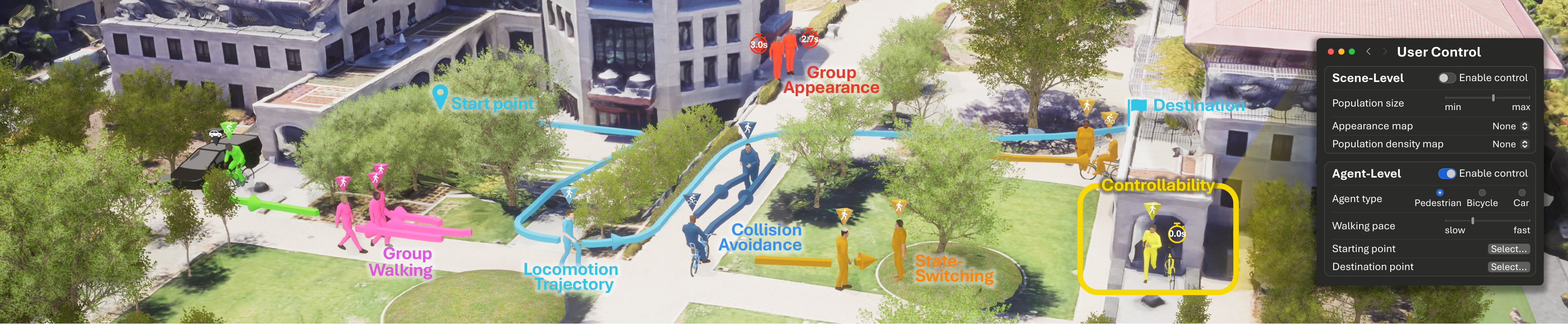}
    \vspace{-7mm}
    \captionof{figure}{Generating realistic, continuous crowd behaviors with learned dynamics. Given a scene image, CrowdES iteratively populates the environment and synthesizes diverse locomotion patterns to create lifelike crowd scenarios. CrowdES also allows users to control parameters to achieve tailored and flexible outcomes.}
    \label{fig:teaser}
\end{center}%
}]
\footnotetext[1]{Corresponding author}
\renewcommand*{\thefootnote}{\arabic{footnote}}

\begin{abstract}
\vspace{-6.5mm}

\noindent%
Modeling and reproducing crowd behaviors are important in various domains including psychology, robotics, transport engineering and virtual environments. Conventional methods have focused on synthesizing momentary scenes, which have difficulty in replicating the continuous nature of real-world crowds. In this paper, we introduce a novel method for automatically generating continuous, realistic crowd trajectories with heterogeneous behaviors and interactions among individuals. We first design a crowd emitter model. To do this, we obtain spatial layouts from single input images, including a segmentation map, appearance map, population density map and population probability, prior to crowd generation. The emitter then continually places individuals on the timeline by assigning independent behavior characteristics such as agents' type, pace, and start/end positions using diffusion models. Next, our crowd simulator produces their long-term locomotions. To simulate diverse actions, it can augment their behaviors based on a Markov chain. As a result, our overall framework populates the scenes with heterogeneous crowd behaviors by alternating between the proposed emitter and simulator. Note that all the components in the proposed framework are user-controllable. Lastly, we propose a benchmark protocol to evaluate the realism and quality of the generated crowds in terms of the scene-level population dynamics and the individual-level trajectory accuracy. We demonstrate that our approach effectively models diverse crowd behavior patterns and generalizes well across different geographical environments. Code is publicly available at \url{https://github.com/InhwanBae/CrowdES}.%
\vspace{-5mm}
\end{abstract}

\section{Introduction}
Crowds exhibit highly complex behaviors, because people are driven by individual goals, and are influenced by other crowd members and environmental factors~\cite{yang2020review}. Synthesizing these behaviors is crucial for various applications such as autonomous driving~\cite{kar2019metasim}, computer games~\cite{unrealsoftware}, virtual cinematography~\cite{reynolds1987flocks} and urban planning~\cite{ucrowdssoftware}.
Unfortunately, creating vivid, lifelike crowd behaviors requires labor-intensive human annotation. Artists or graphic designers must manually position agents in virtual environments, and employ handcrafted intelligence to replicate natural behavioral motions. 

To accelerate these tasks, many efforts have been made to automatically populate the environments using interactive authoring~\cite{ulicny2004crowdbrush} and procedural modeling~\cite{lee2006motion,yersin2009crowd}, and motion synthesis using rule-\cite{reynolds1987flocks}/force-\cite{helbing1995social}/velocity-based approaches~\cite{van2008reciprocal}. Learning-based methods have recently been used to facilitate realistic behavior planning by integrating visual information and contextual cues~\cite{lee2018crowd,rempe2023trace,panayiotou2022ccp,charalambous2023greil}. However, because a wide variety of crowd collective patterns emerge from self-organization, as yet a holistic framework for learning and generating realistic human dynamics is, to the best of our knowledge, still unavailable.

Meanwhile, research into automatic actor placement has begun for traffic scene generation~\cite{tan2021scenegen,feng2023trafficgen,ding2023realgen}. In this task, generative models augment plausible agent layouts on high-definition maps of the surrounding area~\cite{sun2024drivescenegen,lu2024scenecontrol,pronovost2023scenario}. However, by generating momentary 2D distributions of vehicles, such methods are hardly able to handle the continuous, evolving nature of real-world environments.

In this paper, our goal is to automatically produce continuous and realistic crowd trajectories with heterogeneous behaviors and interactions in space from single input images. To achieve this, we introduce CrowdES, a framework consisting of two models: the crowd emitter and simulator. 

\textbf{The crowd emitter}, inspired by particle systems~\cite{bouvier1997crowd}, generates individuals who are characterized by various attributes, and continuously populates the scene over time. The emitter firstly analyzes the input image and then predicts spatial layouts, including a semantic segmentation map, appearance map, population density map and population probability. Here, the appearance map indicates potential locations where people might appear, (\eg, building entrances, scene boundaries) and the population map assesses the crowded areas where people are likely to gather. The population probability computes categorical distribution based on the number of agents in the scene. The number of agents to appear within a certain time-duration are determined based on the distribution. Next, the emitter generates new agents conditioned on the current crowd distribution and the predicted spatial layout. For this, a diffusion model is adopted to produce parameters for each agent, including the agents' attributes, starting and destination coordinates, paces, and appearance times. This allows the timeline of each agent to be planned, where individuals appear at their starting points within a specific time window.

\textbf{The crowd simulator}, inspired by locomotion simulation, produces the trajectory from each agent's starting to end-points. Unlike conventional methods, our simulator enables intermediate behavior augmentations, such as stopping to chat with other agents and avoiding obstacles by moving left or right. Motivated by the Switching Dynamical System (SDS), we define various behavior patterns of the crowd as behavioral modes. Using encoded features for the agent attributes, past behaviors and interactions with neighboring individuals, our simulator computes the transition probabilities between the behavioral modes. The mode is randomly sampled under a Markov chain and fed into the decoder to make decisions about the agent's next steps. Our crowd simulator generates the full route toward the destination through the recurrent prediction process. Once an agent reaches its destination, it disappears from the scene.

Our CrowdES can produce lifelong crowd animations by alternating between the crowd emitter and simulator. Additionally, it allows users to control intermediate outputs with flexibility to customize the population density, starting and destination positions, actor types and walking speed.

Lastly, we build a \textbf{benchmark protocol} to evaluate the performance of our CrowdES and relevant works. Inspired by trajectory prediction tasks, we compare the generated result to ground-truth crowd tracks considering two aspects: (1) assessing realism at the scene level based on crowd distribution and (2) evaluating the accuracy of individual agent trajectories. This benchmark is performed with up to 10 hours of video. Our framework, which benefits from the synergy between the emitter and simulator, can generate realistic crowd animations, even in previously unseen environments. Furthermore, the flexibility of our framework allows us to achieve controllable crowd scenarios.

\section{Related Works}
\subsection{Multi-Agent Trajectory Prediction}
Starting with physical formulation-based methods~\cite{helbing1995social,pellegrini2009you,mehram2009socialforcemodel,yamaguchi2011you}, trajectory prediction models have significantly improved by incorporating neural networks and learning techniques. To infer socially-acceptable paths, many considerations have been implemented for both the agents' interaction and dynamics modeling. Interactions with neighbors are crucial, especially to adhere to social norms such as collision avoidance and group movements. One pioneering work, Social-LSTM~\cite{alahi2016social}, implicitly models the social relations of agents by integrating hidden states for their neighbors through a social pooling module. Subsequent research weighs the mutual influences among agents using attention mechanisms~\cite{vemula2018social,ivanovic2019trajectron,fernando2018soft,salzmann2020trajectron++}, graph convolutional networks~\cite{kipf2016semi,mohamed2020social,sun2020rsbg,bae2021dmrgcn,liu2021snce,liu2021causal,mohamed2022socialimplicit,kim2024higher}, graph attention networks~\cite{velivckovic2018graph,huang2019stgat,liang2020garden,liang2020simaug,bae2022npsn,jeong2024multi,shi2023representing,xu2023eqmotion,tang2024hpnet,wen2024density}, and transformers \cite{yu2020spatio,yuan2021agent,monti2022stt,bae2022gpgraph,wen2022socialode,wong2022v2net,shi2023trajectory,xu2024adapting,wong2024socialcircle,pourkeshavarz2024cadet,fujii2024realtraj,tsao2022socialssl,wong2023another,xia2022cscnet,wong2023msn}. Incorporating additional visual data allows the trajectory prediction models to leverage semantic information about the walkable terrain \cite{varshneya2017,xue2018sslstm,manh2018scene,sadeghian2019sophie,liang2019peeking,kosaraju2019social,zhao2019matf,tao2020dynamic,marchetti2020mantra,marchetti2020multiple,deo2020trajectory,sun2022human,shafiee2021Introvert,mangalam2021ynet,yue2022nsf,wang2023fend,zhang2024oostraj,pourkeshavarz2024dyset,feng2024unitraj,fan2024risk,casas2024detra,thakkar2024adaptive,wangtrajprompt,park2023long,zhao2020tnt,choi2023r,mao2023leapfrog,aydemir2023adapt,chen2023traj,pourkeshavarz2023learn,jiao2023semi,yan2023int2}.

The captured interaction features are then used for dynamics models to predict feasible future trajectories. Predictors adopt either recurrent methods~\cite{rehder2015goal,rehder2018pedestrian,alahi2016social,bisagno2018group,pfeiffer2018,zhang2019srlstm,xu2020cflstm,ma2020autotrajectory,zhao2021experttraj,marchetti2022smemo,navarro2022social,chen2023unsupervised,maeda2023fast,park2024improving,dendorfer2020goalgan,xu2023uncovering,das2024distilling,zhang2020social,dong2023sparse,zhang2023trajpac}, which account for the temporal characteristics of trajectory coordinates, or simultaneous methods \cite{mohamed2020social,bae2021dmrgcn,Shi2021sgcn,pang2021lbebm,li2021stcnet,xu2022remember,bae2023graphtern,lin2024progressive,lee2024mart,gao2024multi,hug2020bezier}, which regress all coordinates at once. Recent works have introduced probabilistic inferences to explain the indeterminacy inherent in crowd behaviors \cite{gupta2018social}. Techniques such as bivariate Gaussian distributions \cite{bae2021dmrgcn,mohamed2020social,shi2020multimodal,li2020Evolvegraph,shi2021socialdpf,yao2021bitrap,Shi2021sgcn,xu2022tgnn,moon2024visiontrap}, generative adversarial networks \cite{gupta2018social,sun2020reciprocal,li2019idl,liang2021tpnms,dendorfer2021mggan,huang2019stgat,sun2023stimulus}, conditional variational autoencoders \cite{lee2017desire,li2019conditional,bhattacharyya2020conditional,mangalam2020pecnet,chen2021disdis,sun2021pccsnet,lee2022musevae,wang2022stepwise,xu2022groupnet,xu2022socialvae,park2024t4p,park2023leveraging}, diffusion models \cite{gu2022mid,mao2023leapfrog,rempe2023trace,jiang2023motiondiffuser,wang2024optimizing}, language models \cite{seff2023motionlm,philion2024trajeglish,bae2024lmtrajectory}, and explicit modeling \cite{kothari2021interpretable,bae2023eigentrajectory,bae2024singulartrajectory,shi2022social} have been used for stochastic trajectory prediction.

Although these trajectory prediction models can generate realistic and diverse behaviors, they face several challenges when attempting unconditional generation, because of their dependency on past trajectories. Additionally, because they use fixed time windows for both the observation and inference steps, it is difficult to plan long-term paths.

\subsection{Crowd Locomotion Simulation}
Boids algorithms~\cite{reynolds1987flocks,reynolds1999steering}, one of the earliest crowd simulation systems, introduce simple rules for alignment, cohesion, and separation to model group dynamics. Since then, crowd simulations have been studied to leverage various group behaviors~\cite{ren2017group}, for collision avoidance~\cite{van2008reciprocal,van2010optimal}, and user-defined constraints and goals~\cite{pettre2008populate,durupinar2009ocean,guy2011simulating,kim2012interactive}. Data-driven methods, which first create databases of example behaviors and then match~\cite{lee2007group,lerner2007crowdsbyexample} or blend~\cite{ju2010morphable,lerner2010context} agents' actions to them during simulations, have further enhanced the diversity and realism of output actions~\cite{lai2005group,kwon2008group,zhao2018clust}. These approaches have evolved to the development of learnable approaches~\cite{lee2018crowd,rempe2023trace,panayiotou2022ccp,yao2020learning,chen2024social}. For instance, GREIL-Crowds~\cite{charalambous2023greil} trains a goal-seeking behavior within a group using guided reinforcement learning. Additionally, it shows continuous crowd simulations by leveraging real-world data, including individual entrance times, origins and goals. However, research on methods to continuously populate scenes with crowds beyond just locomotion, remains limited.

Populating spaces with crowds that exhibit natural behaviors is also critical for broader applications. Commercial software for 3D animation~\cite{houdinisoftware}, visual effects~\cite{golaemsoftware}, urban planning~\cite{ucrowdssoftware} and games~\cite{unrealsoftware} employ functions which treat virtual crowds as particles, in which particle emitter systems randomly generate actors at predefined regions. In this paper, we shift this paradigm into a learnable method to plan how densely to populate scenes with crowds and perform crowd emissions using diffusion models.

\subsection{Traffic Scene Generation}
Recent advances in generative models have enabled the synthesis of realistic traffic scenes~\cite{tan2021scenegen,kar2019metasim,devaranjan2020metasim2,tan2023language,lu2024scenecontrol}. This task generates the initial position, direction and size of vehicles in a scene,  given a map image. Specifically, SceneGen~\cite{tan2021scenegen} employs an LSTM module to autoregressively generate traffic scenes by sequentially inserting actors one at a time. TrafficGen~\cite{feng2023trafficgen} uses an encoder-decoder architecture to sample vehicles' initial states in probability distributions. RealGen~\cite{ding2023realgen} synthesizes traffic by fusing retrieved examples from external data. More recently, several methods have been proposed to leverage the powerful generative abilities of diffusion models for vehicle placement~\cite{sun2024drivescenegen,pronovost2023generating,lu2024scenecontrol,pronovost2023scenario}. Although these works have demonstrated the potential ability to learn to populate environments, they still only focus on agent placement during the initial scene setup. As a result, it is difficult to account for vehicles that enter later. So, once all of the vehicles have left the area of interest, the scenes become empty.

In this paper, we present a diffusion model-based crowd emitter which continually populates scenes with dynamic actors, ensuring lifelong crowd animations.

\subsection{Switching Dynamical Systems}
A dynamical system is a framework based on a set of rules or equations~\cite{bar1996estimation,roweis1999unifying}. For complex dynamical systems, dividing their behaviors into distinct modes, each with simpler dynamics, is often effective~\cite{liberzon2003switching}. Switching Dynamical Systems (SDS) facilitates the identification of these modes and the transitions between them on time series data~\cite{ackerson1970state,ghahramani2000variational,oh2005variational,kurle2020deep,linderman2016recurrent,fraccaro2017disentangled,balsells2024identifying}. In particular, SNLDS~\cite{dong2020collapsed} learns to switch between the discrete states of nonlinear dynamical models. REDSDS~\cite{ansari2021deep} introduces a recurrent state-to-switch connection, along with explicit state duration models, to efficiently capture the duration of varying states. GRASS~\cite{liu2023graph} further advances this approach by employing a dynamic graph-based aggregation to model interaction-aware mode switching. We incorporate SDS into crowd dynamics modeling to better generate agents' dynamic, long-term movement and the behaviors of crowds.

\begin{figure*}[t]
\centering
\includegraphics[width=\linewidth,trim={0 11.8mm 0 0},clip]{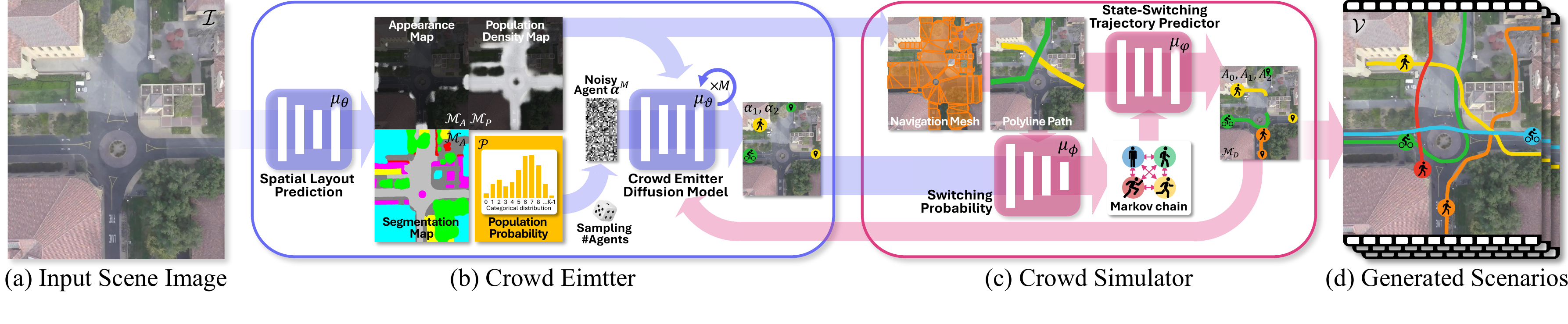}
\vspace{-7mm}
\caption{An overview of our CrowdES framework. Starting with the input scene image $\mathcal{I}$, CrowdES continuously generates realistic crowd behaviors $\mathcal{V}$ by alternating between the crowd emitter and crowd simulator processes.}
\label{fig:method_framework}
\vspace{-2mm}
\end{figure*}

\vspace{-0.5mm}
\section{Methodology}
\vspace{-0.5mm}
We describe a method for modeling the continuous dynamics of crowd behaviors. First, we define the crowd behavior generation problem in~\cref{sec:method_definition}. Next, in~\cref{sec:method_emitter}, we introduce our crowd emitter model to populate environments made from single input images. We then present our crowd simulator model, which produces trajectories from the starting points to the destinations of each agent with intermediate behavior augmentations, in~\cref{sec:method_simulator}. Finally, in~\cref{sec:method_generation}, we integrate both the crowd emitter and crowd simulator within the proposed CrowdES framework. An overview of our CrowdES framework is illustrated in~\cref{fig:method_framework}.

\subsection{Problem Definition}\label{sec:method_definition}
Our goal is to predict realistic trajectories of crowds over time based on single input images. Specifically, given a scene image $\mathcal{I}$, we aim to generate a crowd behavior scenario $\mathcal{V}$ of length $T_\mathcal{V}$ with $N$ agents in total. Each agent $A=\{\kappa, \mathcal{T}\}$ is characterized by an agent type $\kappa$ and a trajectory $\mathcal{T} = [\bm{c}_{T_{\nsp s}}, ...,  \bm{c}_{T_{\nsp d}}]$, where $\bm{c}_t$ represents the 2D coordinate $(x, y)$ at time $t$. Here, $T_{\nsp s}$ and $T_{\nsp d}$ denote the start and destination times for each agent, respectively.

Modeling long and multiple trajectories is challenging because future crowd behaviors are highly correlated with past scene states and interactions with each other. To alleviate this complexity, we adopt an approach that generates crowd behaviors incrementally within a smaller time window of length $T_{\nsp w}$. Within each time window, we capture the emerging and locomotion characteristics of individual agents using a two-stage modeling approach: the crowd emitter and the crowd simulator.

\subsection{Crowd Emitter Model}\label{sec:method_emitter}
To understand terrain geometry and to ensure the controllability of the crowd emitter during inference, we design a preprocessing step that assesses spatial layouts from input scene images. The diffusion model for generating populations then uses the spatial layouts as conditions in a time-sequential manner.

\vspace{1mm}\noindent\textbf{Spatial layout prediction.}\quad
We extract a spatial layout from input scene images $\mathcal{I}$, consisting of a semantic segmentation map $\mathcal{M}_{\nsp S}$, an appearance map $\mathcal{M}_{\nsp A}$, a population density map $\mathcal{M}_{\nsp P}$, and population probability $\mathcal{P}$.

The semantic segmentation maps $\mathcal{M}_{\nsp S}$ categorize each pixel of the scenes into seven classes: buildings, structures, bushes, grasses, trees, sidewalks and roads. The appearance maps $\mathcal{M}_{\nsp A}$ highlight possible locations for people to emerge. To establish the ground-truth of $\mathcal{M}_{\nsp A}$ in a binary image, we gather coordinates $\bm{c}_{T_{\nsp s}}$ and $\bm{c}_{T_{\nsp d}}$ for all $N$ agents, then mark the pixels as 1. The population density maps identify regions where people are likely to come together. We create the $\mathcal{M}_{\nsp P}$ label by counting the number of pixels on a trajectory coordinate $\bm{\tau}$ for all $N$ agents across the image pixels, by applying a logarithmic transformation, and by normalizing the counting values between 0 and 1. The population probabilities $\mathcal{P}$ is a probability made up of $K$ numbers for 0 to $K\!-1$ population, and represent a categorical distribution of how many agents will exist in the current scene. The ground-truth labels for $\mathcal{P}$ are computed by counting the number of agents present across all $T_\mathcal{V}$ frames.

We compute $\mathcal{M}_{\nsp S}$ using a pretrained Grounded-SAM \cite{ren2024grounded} model. 
We also leverage a pretrained SegFormer~\cite{xie2021segformer} backbone $\mu_\theta$ to obtain the others, a segmentation head to predict\,$\mathcal{M}_{\nsp A}$\,\&\,$\mathcal{M}_{\nsp P}$,\,and a classification\,head\,to\,estimate\,$\mathcal{P}$\,as:
\begin{equation}
\begin{gathered}
    \mathcal{M}_{\nsp A}, \mathcal{M}_{\nsp P}, \mathcal{P} = \mu_\theta(\mathcal{I}, \mathcal{M}_{\nsp S}).
\end{gathered}
\end{equation}

\vspace{1mm}\noindent\textbf{Crowd emitter diffusion model.}\quad
To populate the scenes, we need to determine the number of agents $N_{t:t+T_{\nsp w}}$ to be assigned over the interval $t$ to $t+T_{\nsp w}$. This value $N_{t:t+T_{\nsp w}}$ is sampled based on the number of agents in the previous frame $N_{t-1}$, and the population probability distribution $\mathcal{P}$, defined by the following equation:
\begin{equation}
\begin{gathered}
    N_{t:t+T_{\nsp w}} = \min(N_{\mathcal{P}} - N_{t-1}, 0) \\
    ~~~\textit{where}~~~ N_{\mathcal{P}} \sim \text{Categorical}(\mathcal{P}).
\end{gathered}
\end{equation}

\setlength{\belowdisplayskip}{5pt} \setlength{\belowdisplayshortskip}{5pt}
\setlength{\abovedisplayskip}{5pt} \setlength{\abovedisplayshortskip}{5pt}

Next, we implement a diffusion model to iteratively generate agents through a denoising process. Each agent is parameterized by $\bm{\alpha} = \{\kappa, \nu, T_{\nsp s}, \bm{c}_{T_{\nsp s}}, \bm{c}_{T_{\nsp d}}\}$, where $\nu$ denotes the agents’ walking pace. The final agent parameters $\alpha^0$ are progressively denoised from an initial Gaussian noise vector $\alpha^M$ over a sequence of $M$ diffusion steps. In the forward diffusion process, noise is gradually added to the data as:
\begin{equation}\label{eq:diffusion_forward}
\begin{gathered}
q(\bm{\alpha}^{1:M} | \bm{\alpha}^0) := \prod_{m=1}^{M} q(\bm{\alpha}^m | \bm{\alpha}^{m-1}) \vspace{-2pt}\\
\quad q(\bm{\alpha}^m | \bm{\alpha}^{m-1}) := \mathcal{N}\big(\bm{\alpha}^m; \sqrt{1-\beta^m} \bm{\alpha}^{m-1}, \beta^m \bm{\mathrm{I}}\big),
\end{gathered}
\end{equation}
where $\beta_m$ is a constant that defines the noise schedule at each step $m$. The reverse denoising process restores $\alpha^M$ to $\alpha^0$ using a learnable network $\mu$ as follows:
\begin{equation}\label{eq:diffusion_reverse}
\begin{gathered}
p_\vartheta(\bm{\alpha}^{0:M}) := p(\bm{\alpha}^M) \prod_{m=1}^{M} p_\vartheta(\bm{\alpha}^{m-1} | \bm{\alpha}^m), \vspace{-2pt}\\
\!\!p_\vartheta(\bm{\alpha}^{m-1} | \bm{\alpha}^m) := \mathcal{N}\big(\bm{\alpha}^{m-1}; \bm{\mu}_\vartheta(\bm{\alpha}^m,m, \bm{C}_{\nsp e}),\beta^m\bm{\mathrm{I}}\big).\!\!
\end{gathered}
\end{equation}
Here, $\bm{C}_{\nsp e} = \{\mathcal{M}_{\nsp S}, \mathcal{M}_{\nsp A}, \mathcal{M}_{\nsp P}, \mathcal{M}_{\nsp D} \}$ represents the environmental conditions guiding the agent parameters to align with terrain constraints. Specifically, $\mathcal{M}_{\nsp D}$ is a binary map indicating individual positions in the previous frame. The diffusion emitter generates clean agent parameters from Gaussian noise by using a reverse denoising process.

In addition, individuals in crowds often form groups, move collectively, and share their destinations~\cite{moussaid2010walking}. To model this collectivity, our model denoises the parameters of all $N_{t:t+T_{\nsp w}}$ agents, $[\bm{\alpha}_1, ..., \bm{\alpha}_{N_{t:t+T_{\nsp w}}}]$, simultaneously. To be specific, the denoising network $\mu_\vartheta$ leverages transformer architectures to propagate features across agents, and then uses cross-attention with encoded environmental conditions to improve the spatial awareness. In the end, the generated agents are positioned along a timeline, and are placed in the scene by the crowd simulator.

\subsection{Crowd Simulator Model}\label{sec:method_simulator}
To generate environment-aware locomotion that allows agents to navigate from their starting points to destinations, even across complex terrains, we employ a navigation mesh and its benefits. Additionally, we design a network to predict agent behavior dynamics, enabling stochastic switching to achieve continuous and diverse crowd behaviors.

\vspace{1mm}\noindent\textbf{Navigation mesh.}\quad
In simple and open spaces, paths to destinations can often be planned directly with minimal complexity. However, in more complex environments, such as a university campus, routes may need to navigate around various obstacles, sometimes requiring U-shaped or maze-like paths. To simulate both cases, we use a navigation mesh, subdividing the walkable spaces into polygonal regions for path planning ~\cite{snook2000simplified,tozour2002building,van2016comparative}, to define an initial path to the destination. 

First, we create a binary traversable map $\mathcal{M}_{\nsp W}$ by using the segmentation map $\mathcal{M}_{\nsp S}$ to exclude non-navigable areas such as buildings, structures, and bushes. We then exploit the Recast method~\cite{recastsoftware,pyrecastsoftware} to generate a traversable mesh. Using the connectivity of polygons within a mesh, we are able to search for a polyline path from the agent $A$'s current position $\bm{c}_t$ to the destination $\bm{c}_{T_{\nsp d}}$. Along this polyline, we designate $\bm{c}_{t,\textit{nav}}$, a control point located at a distance equal to the agent's walking pace $\nu_n$ from $\bm{c}_t$, for navigation.

\vspace{1mm}\noindent\textbf{Switching crowd dynamical systems.}\quad
For state-switching behavior, we predict the transition probabilities of a Markov chain based on historical data. Inspired by anchor-based trajectory prediction methods~\cite{kothari2021interpretable,chai2019multipath,bae2023eigentrajectory}, we define the behavior states in a data-driven manner. Each agent's trajectory $\mathcal{T}$ is segmented into sequences of length $T_{\!f}$, yielding $\mathcal{T}_{t:t+T_{\!f}}$. The parameter $T_{\!f}$ determines how frequently agents can take different behaviors. Using the origin coordinate $\bm{c}_t$ of each segmented trajectory, we calculate $\bm{c}_{t,\textit{nav}}$. We then normalize the segmented trajectory in terms of translation, rotation, and scale using $\bm{c}_{t,\textit{nav}}$. Here, we apply K-means clustering to obtain $B$ cluster centers, each representing one of the $B$ behavioral states. The index of the cluster center closest to $\mathcal{T}_{t:t+T_{\!f}}$ is assigned as the ground-truth behavior state $\bar{b}_{\nsp f}$.

Next, we predict an agent’s future state $b_{\nsp f}$ based on its previous state $b_h$. In particular, the learnable network $\mu_\phi$ takes agent parameters, historical trajectory, neighborhood trajectories and environmental information as conditions $\bm{C}_{\nsp s}$ to predict $b_{\nsp f}$ as follows:
\begin{equation}
\begin{gathered}
    p_\phi(b_{\nsp f} | b_h) := \text{Categorical}\big(b_{\nsp f}; \mu_\phi(b_h, \bm{C}_{\nsp s}), b_h\big)\\
    \!\!\!\text{s.t.}~~~ \bm{C}_{\nsp s}=\{\alpha, \bm{c}_{t,\textit{nav}}, \mathcal{T}_{t-T_{\nsp h}:t}, \mathcal{H}_{t-T_{\nsp h}:t}, \mathcal{M}_{\nsp S}, \mathcal{M}_{\nsp P}\},\!\!\!
\end{gathered}
\end{equation}
where $\mathcal{T}_{t-T_{\nsp h}:t}$ is the historical trajectory of length $T_{\nsp h}$, and $\mathcal{H}_{t-T_{\nsp h}:t} = \{\mathcal{T}_{t-T_{\nsp h}:t}\}_{n=1}^{N_{t-1}}$ is the neighborhood trajectories used to capture social interactions. The sampled behavior state is then fed into the trajectory prediction network.

\vspace{1mm}\noindent\textbf{State-switching crowd simulator.}\quad
Lastly, we generate agent locomotion in a recurrent manner. Using the agent-environment feature set $\bm{C}_{\nsp s}$ and a sampled behavior state $b_{\nsp f}$, our trajectory predictor $\mu_\varphi$ estimates a realistic future trajectory as follows:
\begin{equation}
    \mathcal{T}_{t:t+T_{\!f}} = \mu_\varphi(b_{\nsp f}, \bm{C}_{\nsp s}).
\end{equation}
Through an iterative inference with $\mu_\varphi$, each agent can have a complete sequence of footsteps from the starting point to the destination.

During network training, we use a ground-truth behavior $\bar{b}_{\nsp f}$, instead of the sampled $b_{\nsp f}$. This enables our model to implement a dynamical system where crowd dynamics can be changed at intervals of $T_{\!f}$.

\begin{figure}[t]
\centering
\includegraphics[width=\linewidth,trim={0 20mm 0 0},clip]{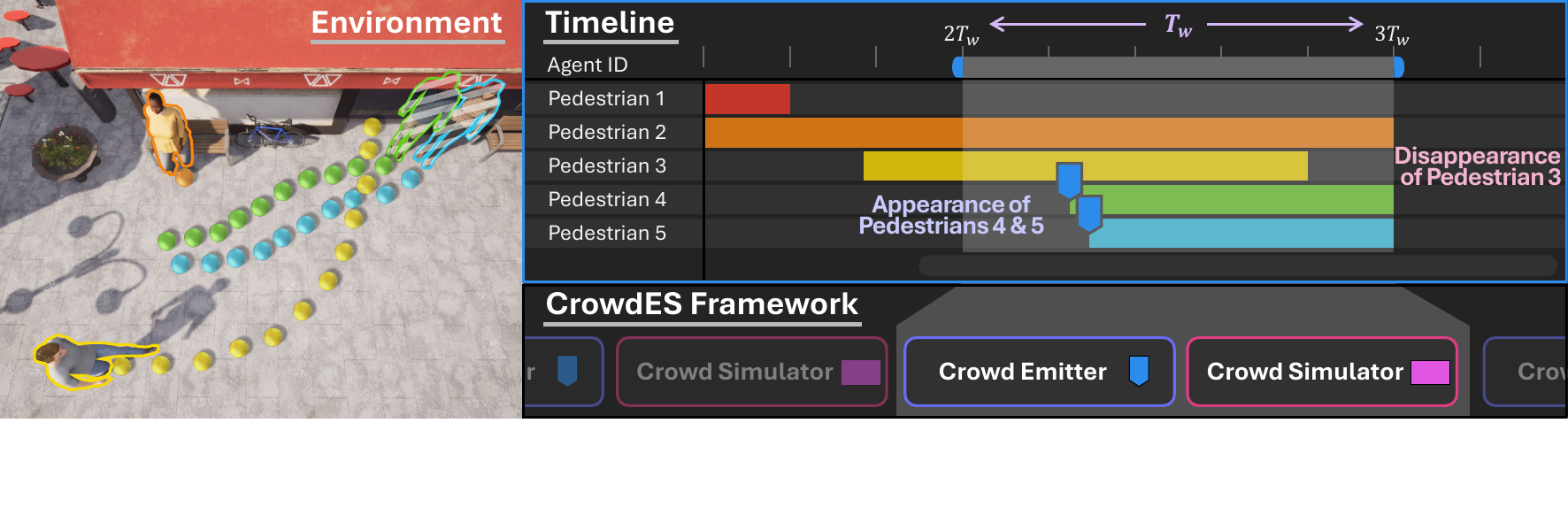}
\vspace{-7mm}
\caption{An overview of our recurrent crowd behavior generation approach using the crowd emitter and simulator. (Marker: Emerging crowds at specific times, Bar: Extensible crowd trajectories.)}
\label{fig:method_timeline}
\vspace{-2mm}
\end{figure}

\subsection{Continuous Crowd Behavior Generation}\label{sec:method_generation}

\vspace{0mm}\noindent\textbf{CrowdES framework.}\quad
Finally, we incorporate the proposed emitter and simulator into our CrowdES framework to generate infinitely long crowd animations over time. For this, we iteratively generate crowd behaviors with a length $T_{\nsp w}$ by alternating between the crowd emitter and simulator, as illustrated in~\cref{fig:method_timeline}. The crowd emitter places the agent markers on the timeline to indicate when agents will emerge over the next $T_{\nsp w}$ frames. The crowd simulator then generates trajectories within this window for all agents in the scene. By repeating this emission-simulation process, our CrowdES produces continuous crowd behaviors, given the environment.

In real-world video footage, people obviously exist throughout the entire space. However, agents in our framework appear one-by-one at the first iteration. To address this, we employ a simple trick: we only populate a half of the total agents $N_{0:T_{\nsp w}}$ during the iteration from 0 to $T_{\nsp w}\!-\!1$, and use a frame at the iteration $T_{\nsp w}$ as an initial frame.

\vspace{1mm}\noindent\textbf{Implementation details.}\quad
We empirically set $T_{\nsp w}$ to 50 frames with a 5 fps system. For the crowd emitter model, we use $M\!=\!50$ diffusion steps and a DDIM scheduler~\cite{song2020ddim} for training. To train $\mu_\theta$, we use a binary cross entropy (BCE) as a loss function between the predicted logits with ground truth maps and probabilities. For $\mu_\vartheta$, we employ a mean square error (MSE) between the output and Gaussian noise.

In the crowd simulation, we set $B\!=\!8$ behavior states, $T_{\nsp h}\!=\!10$, and $T_{\!f}\!=\!20$ in order to prevent frequent behavior switching and ensure smooth motion transitions during simulation. For agent-centric predictions, we crop a map with 16$\times$16 meters centered on each agent's coordinate. To train $\mu_\phi$, we exploit a cross entropy (CE) loss between the estimated and its ground truth behavior states. For $\mu_\varphi$, we use a mean absolute error (MAE) as a loss function between the predicted trajectory and the actual data.

The training is performed with the AdamW optimizer~\cite{loshchilov2018decoupled} at a learning rate of 0.0001. The batch size/training epochs are set to 512/256 and 2048/64 for the crowd emitter and the crowd simulator, respectively. All experiments are conducted on an NVIDIA RTX 4090 GPU, typically taking about one day for each model.

\begin{table}[t]
\centering
\resizebox{\linewidth}{!}{%
\begin{tabular}{@{~~}c@{~~~}c c@{~~~}c@{~~~}c@{~~~}c c@{~~~}c@{~~~}c@{~~~}c}
\toprule
\multirow{2}{*}{Dataset\vspace{-2pt}} & \multirow{2}{*}{Model\vspace{-2pt}} & \multicolumn{4}{c}{Scene-Level Realism} & \multicolumn{4}{c}{Agent-Level Accuracy} \\ \cmidrule(lr){3-6} \cmidrule(lr){7-10}
                       &               & Dens.       & Freq.       & Cov.        & Pop.        & \!\!Kinem.\!\! & DTW      & Div.        & Col.        \\ \midrule
\multirow{3}{*}{ETH}   & SE-ORCA       & 0.054       & 0.034       & 0.034       & 0.640       & \tul{0.502} & \tul{2.122} & 0.188       & \tbf{0.054} \\
                       & VAE           & \tul{0.032} & \tul{0.025} & \tul{0.025} & \tul{0.351} & 1.621       & 3.619       & \tbf{0.285} & 4.959       \\
                       & \tbf{CrowdES} & \tbf{0.020} & \tbf{0.020} & \tbf{0.020} & \tbf{0.208} & \tbf{0.377} & \tbf{1.649} & \tul{0.203} & \tul{0.697} \\ \cmidrule(lr){1-10}
\multirow{3}{*}{HOTEL} & SE-ORCA       & \tul{0.083} & \tul{0.065} & \tul{0.065} & \tul{0.746} & 0.458       & \tul{0.648} & \tbf{0.248} & \tbf{0.045} \\
                       & VAE           & 0.151       & 0.110       & 0.110       & 1.353       & \tbf{0.315} & 0.780       & 0.217       & 1.807       \\
                       & \tbf{CrowdES} & \tbf{0.013} & \tbf{0.009} & \tbf{0.009} & \tbf{0.117} & \tul{0.336} & \tbf{0.643} & \tul{0.242} & \tul{1.197} \\ \cmidrule(lr){1-10}
\multirow{3}{*}{UNIV}  & SE-ORCA       & 0.425       & 0.273       & 0.273       & 0.901       & 0.529       & \tul{1.279} & 0.334       & \tbf{0.011} \\
                       & VAE           & \tul{0.416} & \tul{0.263} & \tul{0.263} & \tul{0.879} & \tul{0.465} & 1.332       & \tbf{0.402} & \tul{0.460} \\
                       & \tbf{CrowdES} & \tbf{0.347} & \tbf{0.204} & \tbf{0.204} & \tbf{0.734} & \tbf{0.420} & \tbf{1.121} & \tul{0.340} & 0.645       \\ \cmidrule(lr){1-10}
\multirow{3}{*}{ZARA1} & SE-ORCA       & 0.025       & 0.018       & 0.018       & 0.389       & \tul{0.473} & \tul{1.256} & \tul{0.227} & \tbf{0.014} \\
                       & VAE           & \tbf{0.014} & \tbf{0.008} & \tbf{0.008} & \tbf{0.231} & 0.576       & 1.342       & 0.221       & \tul{0.827} \\
                       & \tbf{CrowdES} & \tul{0.018} & \tul{0.017} & \tul{0.017} & \tul{0.254} & \tbf{0.327} & \tbf{0.675} & \tbf{0.304} & 1.018       \\ \cmidrule(lr){1-10}
\multirow{3}{*}{ZARA2} & SE-ORCA       & 0.063       & 0.039       & 0.039       & 0.674       & 0.515       & \tul{1.244} & 0.218       & \tbf{0.000} \\
                       & VAE           & \tul{0.049} & \tul{0.026} & \tul{0.026} & \tul{0.534} & \tul{0.473} & 1.309       & \tul{0.244} & \tul{0.630} \\
                       & \tbf{CrowdES} & \tbf{0.009} & \tbf{0.013} & \tbf{0.013} & \tbf{0.100} & \tbf{0.227} & \tbf{0.579} & \tbf{0.355} & 1.021       \\ \cmidrule(lr){1-10}
\multirow{3}{*}{SDD}   & SE-ORCA       & 0.058       & \tul{0.047} & \tul{0.044} & \tul{0.540} & \tul{0.893} & \tul{6.645} & \tul{0.377} & \tbf{0.048} \\
                       & VAE           & \tul{0.052} & 0.052       & 0.047       & 0.678       & 1.078       & 7.881       & \tbf{0.387} & 2.470       \\
                       & \tbf{CrowdES} & \tbf{0.038} & \tbf{0.033} & \tbf{0.030} & \tbf{0.463} & \tbf{0.650} & \tbf{6.352} & 0.354       & \tul{0.411} \\ \cmidrule(lr){1-10}
\multirow{3}{*}{GCS}   & SE-ORCA       & 1.441       & \tbf{0.066} & \tbf{0.066} & 0.617       & \tul{3.419} & \tul{3.289} & 0.197       & \tbf{0.316} \\
                       & VAE           & \tul{0.826} & 0.085       & 0.085       & \tul{0.433} & 5.529       & 5.351       & \tul{0.227} & 9.025       \\
                       & \tbf{CrowdES} & \tbf{0.584} & \tul{0.066} & \tul{0.066} & \tbf{0.329} & \tbf{1.341} & \tbf{3.864} & \tbf{0.279} & \tul{8.519} \\ \cmidrule(lr){1-10}
\multirow{3}{*}{EDIN}  & SE-ORCA       & \tul{0.016} & \tul{0.015} & \tul{0.015} & \tul{0.522} & \tul{2.142} & \tul{1.819} & 0.221       & \tul{0.000} \\
                       & VAE           & 0.031       & 0.031       & 0.031       & 1.375       & 4.704       & 2.292       & \tul{0.375} & 0.001       \\
                       & \tbf{CrowdES} & \tbf{0.002} & \tbf{0.002} & \tbf{0.002} & \tbf{0.313} & \tbf{0.471} & \tbf{1.361} & \tbf{0.386} & \tbf{0.000} \\
\bottomrule
\end{tabular}%
}
\vspace{-3mm}
\caption{Comparison of the CrowdES framework with algorithmic and learnable methods. For Div., higher values indicate better performance; for all other metrics, lower values are better. \tbf{Bold}: Best, \tul{Underline}: Second best.}
\label{tab:exp_evaluation}
\vspace{-2.5mm}
\end{table}

\vspace{-1mm}
\section{Experiments}\label{sec:experiment}
\vspace{-0.5mm}
In this section, we conduct comprehensive experiments to verify the effectiveness of our CrowdES framework. We first describe our benchmark setup in~\cref{sec:exp_benchmark}. Next, we present comparison results with various baseline models on real-world scenes in~\cref{sec:exp_comparison}. We then assess the flexibility and controllability of our framework in both challenging real-world and synthetic environments in~\cref{sec:exp_visualization}. Finally, we perform extensive ablation studies to demonstrate the effects of each component of our framework in~\cref{sec:exp_ablation}.

\vspace{-0.25mm}
\subsection{Benchmark Method}\label{sec:exp_benchmark}
\vspace{-0.25mm}
\vspace{0mm}\noindent\textbf{Datasets.}\quad
To evaluate the realism of the generated crowd scenarios, we used five datasets, including ETH\,\cite{pellegrini2009you}, UCY \cite{lerner2007crowdsbyexample}, Stanford Drone Dataset (SDD)\,\cite{robicquet2016learning}, Grand Central Station dataset (GCS)\,\cite{zhou2012understanding,yi2015understanding}, and Edinburgh dataset (EDIN)\,\cite{majecka2009statistical}. 

The ETH-UCY datasets consist of five subsets, containing a total of 2,329 pedestrians recorded for over one hour of surveillance video. We re-label the dataset to address issues with fragmented and missing trajectory segments. Following \cite{gupta2018social}, we adopt a leave-one-out strategy for both training and evaluation. 
The SDD dataset includes 10,065 agents across six categories (pedestrians, bicyclists, skateboarders, cars, carts and buses). In total, there are 60 video clips, for about 5 hours, captured on a university campus. Following~\cite{kothari2021trajnet}, 17 scenes are used for evaluation, while the remainder are used for training. For ETH-UCY and SDD, their test splits are the unseen environments over training sets.

The GCS dataset provides 1.11 hours of video taken in a highly congested terminal capturing 13,394 pedestrians. For evaluation, we divide the video into an 80$\%$-20$\%$ train-test split.
The EDIN dataset consists of 118 video clips, 873 hours and tracking 108,993 pedestrians, captured at another university campus. We use the clips recorded in December for evaluation, while the remainder are utilized for training.

\begin{figure}[t]
\centering
\includegraphics[width=\linewidth]{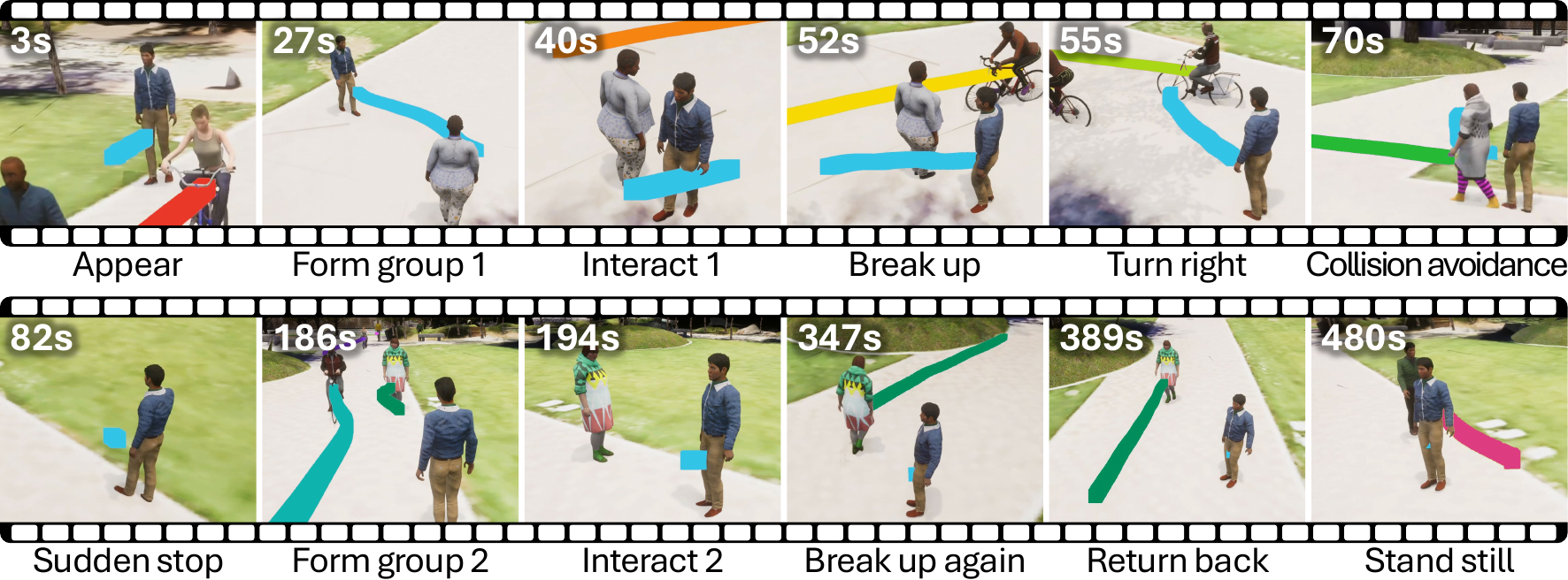}
\vspace{-7mm}
\caption{Visualization of the time-varying behavior changes (blue man). Our CrowdES can autonomously generate realistic, long-term behavioral sequences for each agent in a scene.}
\label{fig:exp_behaviorchange}
\vspace{-3mm}
\end{figure}

\begin{figure}[t]
\centering
\includegraphics[width=\linewidth,trim={0 39.5mm 0 0},clip]{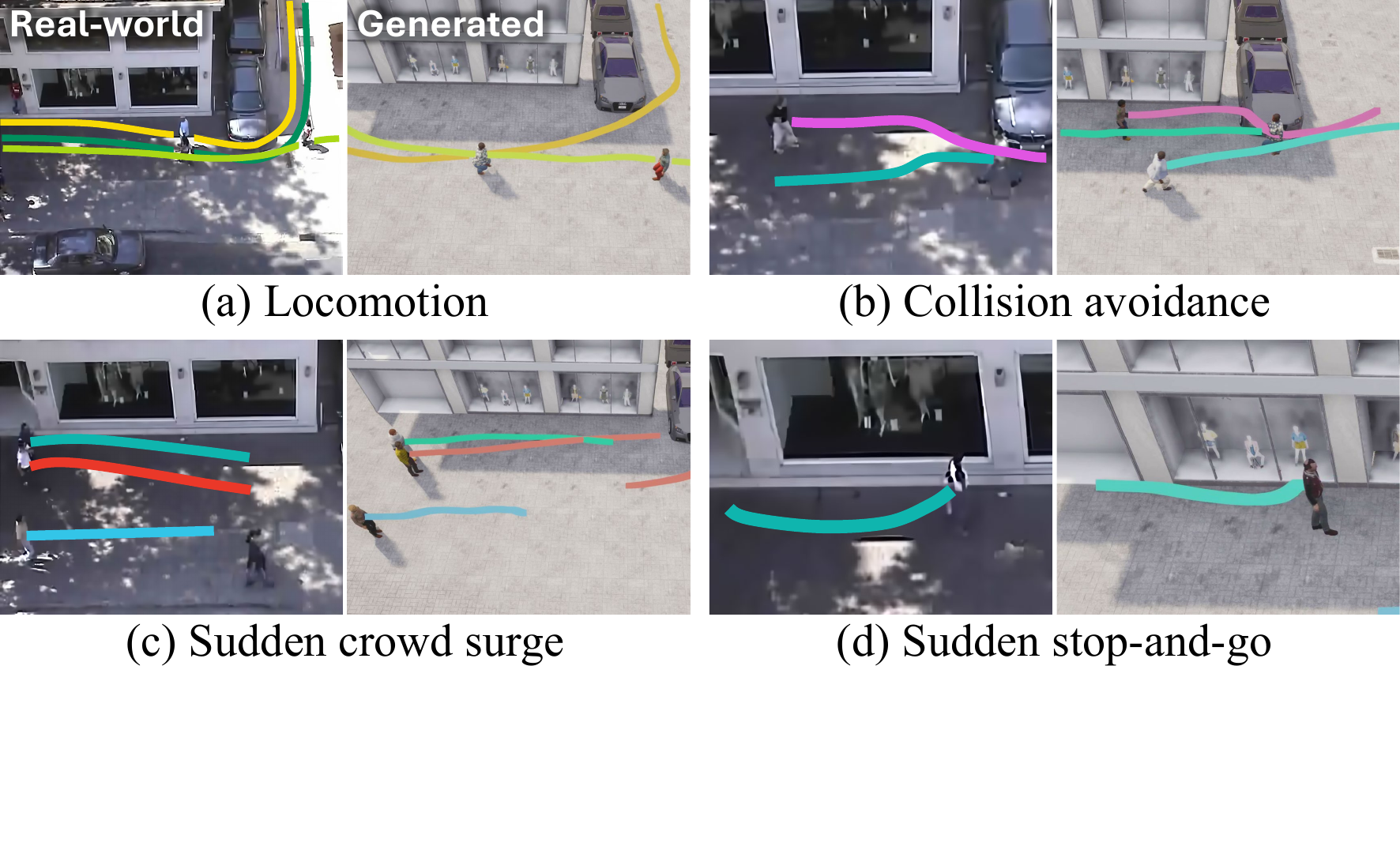}
\vspace{-7mm}
\caption{Visualization of the generated behaviors compared to the real-world behaviors in the ZARA1 scene.}
\label{fig:exp_visualization}
\vspace{-3mm}
\end{figure}

\vspace{1mm}\noindent\textbf{Evaluation metrics.}\quad
There have been various methodologies proposed to assess crowd behaviors, including entropy-based similarity metrics~\cite{guy2012statistical, karamouzas2018crowd}, agent-level accuracy, realism, and certainty metrics~\cite{sohn2022a2x}, and group frequency and density metrics~\cite{o2019assessing}. These approaches inherently require a shared observation state and manually annotated group labels, making them unsuitable for our framework. Meanwhile, Itatani \etal~\cite{itatani2024social} suggest a user-study-based realism measure, which is impractical for evaluating our long videos. To address these limitations, we introduce eight evaluation metrics designed for our crowd behavior generation benchmark, incorporating key insights from previous studies.

In order to evaluate scene-level similarity, we define realism metrics based on the Earth Mover's Distance (EMD)~\cite{rubner1998metric}. Inspired by the quadrat sampling method~\cite{pound1898ii}, we measure the distribution similarity of crowds with respect to the agent types. Specifically, we calculate density (Dens.), frequency (Freq.), and coverage (Cov.) every second on a 10$\times$10 grid for both the generated and ground-truth test sets, and then compute the EMD between them. Similarly, we assess population similarity (Pop.) between the generated and ground-truth histograms of agent counts, collected every second.

To estimate the agent-level similarity, we introduce a kinematics (Kinem.) metric, which averages four EMDs of travel velocities, accelerations, distances and times for all agents in the generated and ground-truth sets. To measure the spatial accuracy of trajectories at meter scale, we calculate the average minimum pairwise Dynamic Time Warping (DTW) distance~\cite{salvador2007toward} between the generated and ground-truth trajectories. Diversity (Div.) is evaluated by examining how comprehensively these minimum-distance pairs cover the full set of agent trajectories. Lastly, we use collision rate (Col.) to check the percentage of generated cases where agents collide with each other.

\vspace{1mm}\noindent\textbf{Evaluation methodology.}\quad
Starting from scene images in the test datasets, the comparison models are intended to generate scenarios composed of multiple crowd trajectories. To evaluate the metrics, the generated scenarios are truncated to have the same duration as the ground-truth video. For a fair comparison, each experiment is repeated 20 times, and the results are averaged to reduce randomness and enhance numerical stability. Details are in our \textit{supplementary\,material}.

\begin{figure}[t]
\centering
\includegraphics[width=\linewidth]{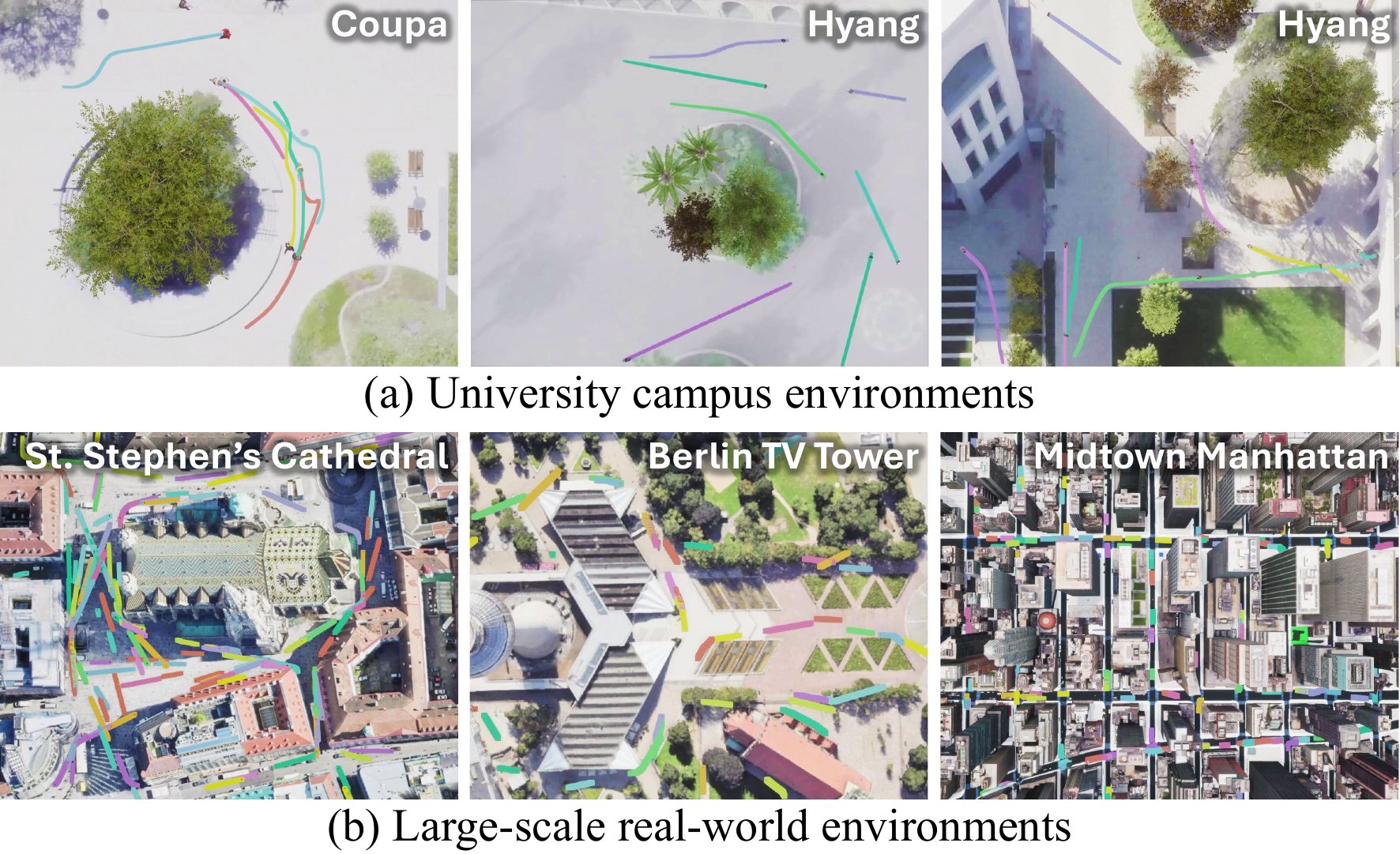}
\vspace{-7mm}
\caption{Visualization of the flexibility of our CrowdES framework across diverse university campuses and large-scale real-world environments (Line: 5-second future trajectories).}
\label{fig:exp_flexibility}
\vspace{-2mm}
\end{figure}

\subsection{Evaluation Results}\label{sec:exp_comparison}
We compare our CrowdES framework with two baseline models, including algorithmic and learnable methods. SE-ORCA combines a random surface emitter~\cite{houdinisoftware} and ORCA~\cite{van2010optimal} algorithms to implement the continuous crowd emerging and collision-avoidance actions. VAE~\cite{mangalam2020pecnet} uses a conditional variational auto-encoder, in place of the conditional diffusion model in the crowd emitter, and SDS in the crowd simulator, to learn the distributions of crowd behavior dynamics. 

In~\cref{tab:exp_evaluation}, we report the comparison results with respect to both scene-level realism and agent-level accuracy. The results demonstrate that our CrowdES consistently outperforms these methods across nearly all metrics and datasets. In particular, our diffusion-based crowd emitter successfully populates the environments with a highly realistic distribution of agents while facilitating group formation. At the same time, the generated agents exhibit realistic kinematics and follow trajectories that closely resemble actual paths. Additionally, compared to the conventional algorithmic approach \cite{van2010optimal}, the probabilistic behavior-switching mechanism significantly enhances the diversity of the generated paths.
However, CrowdES occasionally encounters a few more collisions than SE-ORCA~\cite{van2010optimal}, focusing on collision avoidance ability. This is because our model concentrates on diverse behavioral switching motions. Nevertheless, since the collisions are reported as percentage units, our crowd simulator, which has neighborhood awareness, typically achieves safe locomotion trajectories. We note that CrowdES generates crowd behaviors in real-time, typically requiring 48 seconds to generate a one-hour scenario.

\begin{figure}[t]
\centering
\includegraphics[width=\linewidth]{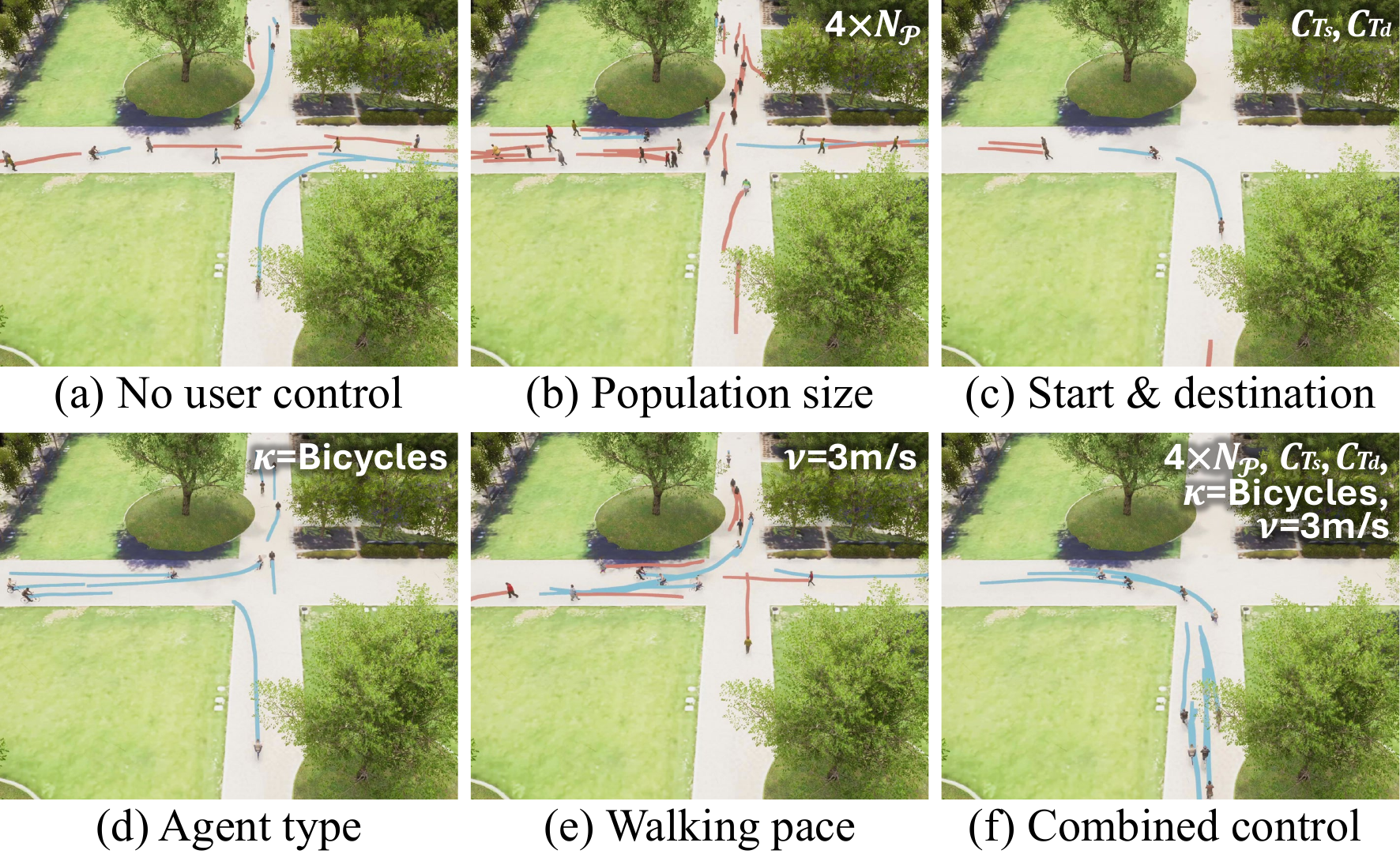}
\vspace{-7mm}
\caption{Visualization of the controllability of our CrowdES frame-work with various user controls (Red: Pedestrians, Blue: Bicycles).}
\label{fig:exp_controllability}
\vspace{-3mm}
\end{figure}

\subsection{Flexibility and Controllability}\label{sec:exp_visualization}
We visualize CrowdES's ability to generate diverse crowd behaviors by presenting our results alongside several real-world cases.
To aid visualization, we synthesize virtual crowds using the CARLA simulator~\cite{dosovitskiy2017carla}.
In~\cref{fig:exp_behaviorchange}, we track an individual throughout the generated long-term scenario to show the socially acceptable behavioral diversity. Our CrowdES framework produces realistic, long-term behaviors with diverse intermediate interactions using probabilistic state transitions within a Markov chain.
In~\cref{fig:exp_visualization}, our crowd simulator successfully produces natural, environment-aware paths from starting points to destinations, which is essential for crowd behavior modeling. Collision avoidance is another key element, and our network successfully learns this behavior in a data-driven manner. Notably, CrowdES is capable of replicating sudden crowd behaviors, such as sudden surges and stop-and-go motions, without any supervision.

Next, we demonstrate the flexibility of our framework to populate crowds and to make their movements within real-world environments in~\cref{fig:exp_flexibility}(a). In a university campus scene, virtual crowds naturally emerge from buildings or scene boundaries and move toward their destinations. During their locomotion toward destinations, crowds avoid collisions with both environmental obstacles like trees and bushes, and other agents.
To emphasize the robustness of our CrowdES, we design three more large-scale and complex real-world environments in~\cref{fig:exp_flexibility}(b). In the Cathedral scene, our crowd simulator generates smooth locomotion paths between buildings. In the TV tower scene, our framework produces realistic crowd behaviors within the intricate structures. Lastly, in the extremely challenging Manhattan scenario, we observe that agents successfully navigate and populate the dense, large-scale urban landscape.

Lastly, we demonstrate the controllability of our CrowdES framework in~\cref{fig:exp_controllability}. Starting from an initial uncontrolled scenario, users can adjust scene-level parameters, including the overall size of population and start/destination areas. At the agent level, users can also edit all agent parameters, for example, changing the agent types from pedestrians to bicycles or accelerating each agent's pace. These components are customizable and can be adjusted simultaneously. As a result, CrowdES not only fully automatically synthesizes crowd behaviors in complex environments, but also offers controllability for various applications.

\begin{table}[t]
\centering
\resizebox{\linewidth}{!}{%
\begin{tabular}{c@{~~~~}l c c@{~~~~}c@{~~~~}c@{~~~~}c}
\toprule
Model & Component & & Dens. & Freq. & Cov. & Pop. \\ \midrule
\multirow{5}{*}{~Crowd Emitter~} & w/o\,\,Diffusion model          & &      0.052  &      0.052  &      0.047  &      0.678  \\
                                 & w/o\,\,Population sampling      & &      0.051  &      0.042  &      0.040  &      0.565  \\
                                 & w/o\,\,Spatial layout condition & & \tul{0.039} &      0.039  &      0.037  &      0.486  \\
                                 & w/o\,\,Collectivity transformer & & \tul{0.039} & \tul{0.037} & \tul{0.033} & \tul{0.478} \\ \cmidrule(r){2-7}
                                 & \tbf{CrowdES}                   & & \tbf{0.038} & \tbf{0.033} & \tbf{0.030} & \tbf{0.463} \\
\bottomrule
\end{tabular}%
}
\vspace{-3.5mm}
\caption{Ablation study of the crowd emitter.}
\label{tab:abl_emitter}
\vspace{-2mm}
\end{table}

\begin{table}[t]
\centering
\resizebox{\linewidth}{!}{%
\begin{tabular}{c@{~~~~}l c c@{~~~~}c@{~~~~}c@{~~~~}c}
\toprule
Model & Component & & \!\!Kinem.\!\! & DTW & Div. & Col. \\ \midrule
\multirow{5}{*}{\!Crowd Simulator\!} & w/o\,\,Spatial layout condition   & & 0.671       & 6.703       & \tul{0.350} & 1.554       \\
                                     & w/o\,\,Navigation mesh            & & \tbf{0.619} & 6.658       & 0.348       & \tul{0.602} \\
                                     & w/o\,\,Social interaction         & & 0.709       & 6.937       & 0.334       & 2.709       \\
                                     & w/o\,\,State-switching            & & 0.706       & \tbf{6.351} & 0.343       & 1.714       \\ \cmidrule(r){2-7}
                                     & \tbf{CrowdES}                     & & \tul{0.650} & \tul{6.352} & \tbf{0.354} & \tbf{0.411} \\
\bottomrule
\end{tabular}%
}
\vspace{-3.5mm}
\caption{Ablation study of the crowd simulator.}
\label{tab:abl_simulator}
\vspace{-2.5mm}
\end{table}

\subsection{Ablation Studies}\label{sec:exp_ablation}

\noindent\textbf{Component of the crowd emitter.}\quad
We conduct an ablation study for the crowd emitter on the SDD dataset by removing its components one-by-one, in~\cref{tab:abl_emitter}. First, we evaluate the backbone generative model. We determine that incorporating the powerful generative capabilities of diffusion models produces a more realistic population distribution compared to VAE models. Next, we replace the population probability prediction with a population regression network. The regression approach fails to capture dynamic population shifts, such as sudden crowd surges, which decreases the scene-level similarities. Third, we remove the spatial layout, conditioned by the diffusion models. We confirm that the appearance and population density maps, commonly used as user controls in conventional software, are also beneficial within a learnable framework. Lastly, we examine the independent agent parameter generation in place of the collective generation. In particular, our collective generation approach improves the performance of the frequency and coverage metrics, which are sensitive to group cohesion.

\noindent\textbf{Components of the crowd simulator.}\quad
Next, we conduct another ablation study of the crowd simulator in~\cref{tab:abl_simulator}. First, the spatial layout and navigation mesh effectively constrain agents to traversable areas, guiding them to plan paths more closely aligned with real trajectories according to the DTW metric. We then remove the social interaction module, which leads to a reduction in diversity and increases collision cases. We also evaluate the state-switching module. The dynamic behavior transitions enable us to simulate unpredictable behaviors, such as sudden stops and stop-and-go motions, which improve both kinematic similarity and behavioral diversity. Lastly, we explore the impact of the number of behavior states for optimal performance. As shown in~\cref{tab:abl_states}, $B\!=\!8$ states achieve the best performance. While additional behavior states may enhance diversity and improve the likelihood of corresponding real trajectories, excessive behavior transitions lead to lower kinematic fidelity.

\begin{table}[t]
\centering
\resizebox{\linewidth}{!}{%
\begin{tabular}{c @{~~}c c@{~~~~}c@{~~~~}c@{~~~~}c c c@{~~~~}c@{~~~~}c@{~~~~}c c}
\toprule
\multirow{2}{*}{\tworow{\#\,Behavior}{States}\vspace{-4pt}} & & \multicolumn{4}{c}{Scene-Level Realism} & \! & \multicolumn{4}{c}{Agent-Level Accuracy} & \!\!\!\! \\ \cmidrule(){3-6} \cmidrule(){8-11}
             & & Dens.       & Freq.       & Cov.        & Pop.        & \! & \!\!Kinem.\!\! & DTW      & Div.        & Col.        \\ \midrule
$B=1$        & & 0.041       & 0.038       & 0.035       & 0.513       & \! & 0.706       & 6.351       & 0.343       & 1.735       \\
$B=2$        & & \tul{0.039} & \tul{0.035} & 0.032       & 0.486       & \! & 0.667       & 6.256       & 0.352       & 1.030       \\
$B=4$        & & \tbf{0.038} & \tbf{0.033} & \tul{0.031} & \tul{0.464} & \! & \tul{0.656} & \tul{6.236} & 0.349       & 0.825       \\
$B=8$        & & \tbf{0.038} & \tbf{0.033} & \tbf{0.030} & \tbf{0.463} & \! & \tbf{0.650} & 6.352       & \tul{0.354} & \tul{0.411} \\
$B=16$\!\!\! & & \tbf{0.038} & \tbf{0.033} & \tul{0.031} & \tul{0.464} & \! & \tul{0.656} & \tul{6.236} & 0.353       & \tbf{0.333} \\
$B=32$\!\!\! & & \tul{0.039} & \tul{0.035} & 0.032       & 0.476       & \! & 0.674       & \tbf{6.225} & \tbf{0.357} & 0.497       \\
\bottomrule
\end{tabular}%
}
\vspace{-3.5mm}
\caption{Ablation study on the number of behavior states.}
\label{tab:abl_states}
\vspace{-2mm}
\end{table}

\section{Conclusion}
In this paper, we present CrowdES, a framework for generating continuous and realistic crowd trajectories with diverse behaviors from single input images. By combining the crowd emitter which assigns individual attributes with the crowd simulator that produces detailed trajectories, our method captures complex interactions and heterogeneity among crowds. Our framework is also user-controllable, allowing customization of parameters such as population density and walking speed. In addition, we introduce a new evaluation protocol for continuous crowd generation tasks. Through a variety of experiments, we demonstrate that our CrowdES generates lifelike crowd behaviors with respect to both scene-level realism and individual trajectory accuracy across diverse environments for dynamic crowd simulation.

\vspace{1.5mm}
\fontsize{7.2}{8}\selectfont{%
\noindent\textbf{Acknowledgement}
This work was supported by the National Research Foundation of Korea (NRF) grant funded by the Korea government (MSIT)(RS-2024-00338439), Institute of Information \& communications Technology Planning \& Evaluation (IITP) grant (RS-2021-II212068, Artificial Intelligence Innovation Hub), and the Korea Agency for Infrastructure Technology Advancement (KAIA) grant funded by the Ministry of Land Infrastructure and Transport (Grant RS-2023-00256888).
}

\normalsize

{
    \small
    \bibliographystyle{ieeenat_fullname}
    \bibliography{egbib}
}

\clearpage
\appendix
\maketitlesupplementary
\normalsize

~

\section{Further Benchmark Details}
In this section, we further describe the details of our benchmark. We first explain the datasets used for our evaluation in~\cref{sec:supp_benchmark_dataset}. We then provide the formulation of the eight evaluation metrics which are proposed to measure the performance of our framework in~\cref{sec:supp_benchmark_metrics}.

\subsection{Datasets}\label{sec:supp_benchmark_dataset}
We carefully evaluate the realism of the generated crowd scenarios using five datasets. Because the datasets composed of short video clips lasting only a few seconds are not enough to capture the continuous nature of crowd behaviors, we employ datasets that track multiple agents with more than 1 minute running time in each location. Specifically, we use the ETH~\cite{pellegrini2009you} and UCY~\cite{lerner2007crowdsbyexample} datasets, which are the most widely used datasets in trajectory prediction tasks. Next, we incorporate the Stanford Drone Dataset (SDD)~\cite{robicquet2016learning} to demonstrate the capability of our framework to handle heterogeneous agent types. Additionally, we employ the Grand Central Station dataset (GCS)~\cite{yi2015understanding}, featuring highly crowded scenes with up to 332 pedestrians simultaneously navigating a train station environment. Lastly, we include the Edinburgh dataset (EDIN)~\cite{majecka2009statistical}, which tracks over 100,000 individuals over a year. The statistics of each dataset used for evaluation are summarized in~\cref{tab:supp_dataset}.

\begin{table}[h]
\vspace{2mm}
\centering
\resizebox{1\linewidth}{!}{%
\begin{tabular}{@{}cc @{~~}c@{~~}c@{~~}c@{~~}c@{~~}c@{~~}c@{~~}c@{~~}c}
\toprule
\multicolumn{2}{c}{Datasets}                  & ~~ETH~~  & HOTEL & ~UNIV~ & ZARA1 & ZARA2 & ~~SDD~~  & ~~GCS~~   & EDIN   \\ \midrule
\multirow{4}{*}{Training\!\!\!\!}   & \#Videos    & 6    & 6     & 5    & 6     & 6     & 43   & 1     & 103    \\ \cmidrule(lr){2-10}
                                    & FPS         & 25   & 25    & 25   & 25    & 25    & 30   & 25   & 10     \\
                                    & Duration(h) & 0.62 & 0.65  & 0.75 & 0.76  & 0.75  & 3.64 & 0.89 & 768.28 \\
                                    & \#Agents    & 1923 & 1754  & 1490 & 2164  & 2122  & 8224 & 9322 & \!101850\! \\ \midrule
\multirow{4}{*}{Evaluation\!\!\!\!} & \#Videos    & 1    & 1     & 2    & 1     & 1     & 17   & 1    & 15     \\ \cmidrule(lr){2-10}
                                    & FPS         & 25   & 25    & 25   & 25    & 25    & 30   & 25   & 10     \\
                                    & Duration(h) & 0.24 & 0.21  & 0.11 & 0.10  & 0.12  & 1.20 & 0.22 & 104.49 \\
                                    & \#Agents    & 406  & 575   & 839  & 165   & 207   & 1841 & 4072 & 7143   \\
\bottomrule
\end{tabular}%
}
\vspace{-3.5mm}
\caption{Dataset statistics used in our benchmark.}
\vspace{-2mm}
\label{tab:supp_dataset}
\end{table}

\subsection{Evaluation Metrics}\label{sec:supp_benchmark_metrics}
We measure the performance of crowd behavior generation models using eight metrics: Density (\textit{Dens.}), Frequency (\textit{Freq.}), Coverage (\textit{Cov.}), Population similarity (\textit{Pop.}), Kinematics (\textit{Kinem.}), Minimum pairwise dynamic time warping distance (\textit{DTW}), Diversity (\textit{Div.}), and Collision rate (\textit{Col.}). Among these, \textit{Dens.}, \textit{Freq.}, \textit{Cov.} and \textit{Pop.} are considered scene-level similarity metrics, while \textit{Kinem.}, \textit{DTW}, \textit{Div.} and \textit{Col.} are classified as agent-level similarity metrics. Given the ground-truth crowd behavior scenario $\mathcal{V}_\textit{GT}$ and the generated scenario $\mathcal{V}_\textit{Gen}$, the eight metrics are defined as follows.

~

\bigskip

\vspace{1mm}\noindent\textbf{Density (\textit{Dens.})}\quad
The similarity between the density distribution of the generated and the ground-truth scenarios. Inspired by the quadrat sampling method~\cite{pound1898ii}, we divide the scene into a $Q\times Q$ grid with $Q=10$. For each time step, the average number of agents per quadrat is computed to estimate the density of the scenario. These values are aggregated over the whole time to construct the density distribution over time. The similarity between the generated and ground-truth distributions is then measured using Earth Mover's Distance (EMD)~\cite{rubner1998metric}. With the indicator function $\mathbb{I}(\cdot)$, which evaluates to 1 if the condition is true and 0 otherwise, the Density (\textit{Dens.}) metric is defined as:
\begin{equation}
\begin{gathered}
    \textit{Dens.}_\textit{EMD} = \textit{EMD}\big( \{\mathcal{D}^\textit{Gen}(t)\}_{t=0}^{T_{\mathcal{V}_\textit{Gen}}\!\!-1}\!, \{\mathcal{D}^\textit{GT}(t)\}_{t=0}^{T_{\mathcal{V}_\textit{GT}}\!-1} \big),\\
    ~~~~~where~~~~~ \mathcal{D}(t) = \frac{1}{Q^2}\sum_{q=1}^{Q^2} \sum_{i=1}^{N_t} \mathbb{I}(\bm{c}_t^i \in q).
\end{gathered}
\end{equation}

\vspace{1mm}\noindent\textbf{Frequency (\textit{Freq.})}\quad
The similarity between the frequency distribution of the generated and ground-truth scenarios. Similar to the \textit{Dens.} metric, the average number of unique agent types per quadrat is computed for each time step to represent the frequency distribution of the scenario. The similarity between the generated and ground-truth frequency distribution is then measured using EMD. Using the cardinality function $|\cdot|$, which counts the number of unique elements, the Frequency (\textit{Freq.}) metric is defined as:
\begin{equation}
\begin{gathered}
    \textit{Freq.}_\textit{EMD} = \textit{EMD}\big( \{\mathcal{F}^\textit{Gen}(t)\}_{t=0}^{T_{\mathcal{V}_\textit{Gen}}\!\!-1}\!, \{\mathcal{F}^\textit{GT}(t)\}_{t=0}^{T_{\mathcal{V}_\textit{GT}}\!-1} \big),\\ 
    ~~~~~where~~~~~ \mathcal{F}(t) = \frac{1}{Q^2}\sum_{q=1}^{Q^2} \big|\{\bm{c}_t^i \in q : \kappa_i\}_{i=1}^{N_t} \big|.
\end{gathered}
\end{equation}

\vspace{1mm}\noindent\textbf{Coverage (\textit{Cov.})}\quad
The similarity between the coverage distribution of the generated and ground-truth scenarios. In a manner similar to the \textit{Dens.} metric, the proportion of quadrats that are occupied at each time step is computed to determine the coverage of the scenario. We then evaluate the similarity of the coverage between the generated and ground-truth distribution using EMD. The Coverage (\textit{Cov.}) metric is defined as:
\begin{equation}
\begin{gathered}
    \textit{Cov.}_\textit{EMD} = \textit{EMD}\big( \{\mathcal{C}^\textit{Gen}(t)\}_{t=0}^{T_{\mathcal{V}_\textit{Gen}}\!\!-1}\!, \{\mathcal{C}^\textit{GT}(t)\}_{t=0}^{T_{\mathcal{V}_\textit{GT}}\!-1} \big),\\ 
    ~~~~~where~~~~~ \mathcal{C}(t) = \frac{1}{Q^2}\sum_{q=1}^{Q^2} \mathbb{I}\Big( \sum_{i=1}^{N_t} \mathbb{I}(\bm{c}_t^i \in q) >0\Big).
\end{gathered}
\end{equation}

\vspace{1mm}\noindent\textbf{Population Similarity (\textit{Pop.})}\quad
The similarity between the population distribution of the generated and ground-truth scenarios. The population size at each time step is aggregated over the scenario duration to create a time-varying population distribution. The EMD is then used to measure the similarity between the generated and ground-truth population distribution. The Population Similarity (\textit{Pop.}) metric is defined as:
\begin{equation}
    \textit{Pop.}_\textit{EMD} = \textit{EMD}\big( \{N_t^\textit{Gen}\}_{t=0}^{T_{\mathcal{V}_\textit{Gen}}\!\!-1}, \{N_t^\textit{GT}\}_{t=0}^{T_{\mathcal{V}_\textit{GT}}\!-1} \big)
\end{equation}

\medskip
\vspace{1mm}\noindent\textbf{Kinematics (\textit{Kinem.}})\quad
The similarity of kinematic properties between the generated and ground-truth scenarios. The kinematic properties consider travel distance, velocity, acceleration and time (duration), measured in metric units. The metric is computed as the average of four EMD measures: (1) $\textit{EMD}_\textit{dist}$ measures the similarity of the total distance traveled by agents, (2) $\textit{EMD}_\textit{val}$ measures the similarity of velocity profiles, (3) $\textit{EMD}_\textit{acc}$ measures the similarity of acceleration profiles, and (4) $\textit{EMD}_\textit{time}$ measures the similarity of travel durations. The overall Kinematics (\textit{Kinem.}) metric is calculated as:
\begin{equation}
\begin{gathered}
    \textit{Kinem.}_\textit{EMD}\!=\!\frac{1}{4} (\textit{EMD}_\textit{dist}\!+\!\textit{EMD}_\textit{val}\!+\!\textit{EMD}_\textit{acc}\!+\!\textit{EMD}_\textit{time}),\medskip\\
    \textit{EMD}_\textit{dist} = \textit{EMD}\big(\{\lVert \mathcal{T}_i^\textit{Gen} \rVert\}_{i=1}^{N^\textit{Gen}}, \{\lVert \mathcal{T}_j^\textit{GT} \rVert\}_{j=1}^{N^\textit{GT}}\big)~~~~~~~~~~~~~~~ \\ ~~~~~~~~~~~~~~~~~~~~~~~ ~~~~~~~where~~~~~~~ \lVert\mathcal{T}\rVert = \sum_{t=T_s}^{T_d-1} \lVert\bm{c}_{t+1} - \bm{c}_{t} \rVert_2, \medskip\\    \textit{EMD}_\textit{vel} = \textit{EMD}\big(\{\lVert\dot{\mathcal{T}}_i^\textit{Gen}\rVert\}_{i=1}^{N^\textit{Gen}}, \{\lVert\dot{\mathcal{T}}_j^\textit{GT}\rVert\}_{j=1}^{N^\textit{GT}}\big)~~~~~~~~~~~~~~~ \\ ~~~~~~~~~~~~~~~~~~~~~~~~~~~~~~~~~~~~~~~~~~~~~~ ~~~~~~~where~~~~~~~ \dot{\mathcal{T}} = \frac{d\mathcal{T}(t)}{dt}, \medskip\\
    \textit{EMD}_\textit{acc} = \textit{EMD}\big(\{\lVert\ddot{\mathcal{T}}_i^\textit{Gen}\rVert\}_{i=1}^{N^\textit{Gen}}, \{\lVert\ddot{\mathcal{T}}_j^\textit{GT}\rVert\}_{j=1}^{N^\textit{GT}}\big)~~~~~~~~~~~~~~~ \\ ~~~~~~~~~~~~~~~~~~~~~~~~~~~~~~~~~~~~~~~~~~~~ ~~~~~~~where~~~~~~~ \ddot{\mathcal{T}} = \frac{d^2\mathcal{T}(t)}{dt^2}, \medskip\\
    \textit{EMD}_\textit{time} = \textit{EMD}\big(\{\tau_{i}^\textit{Gen}\}_{i=1}^{N^\textit{Gen}}, \{\tau_j^\textit{GT}\}_{j=1}^{N^\textit{GT}}\big)~~~~~~~~~~~~~~~~~~~~~~~ \\ ~~~~~~~~~~~~~~~~~~~~~~~~~~~~ ~~~~~~~~ ~~~~~~~where~~~~~~~ \tau_i = T_d-T_s+1.
\end{gathered}
\end{equation}

\medskip
\vspace{1mm}\noindent\textbf{Minimum Pairwise Dynamic Time Warping (\textit{DTW})}\quad
The spatial alignment of trajectories in the generated scenario over the ground-truth scenario using the Dynamic Time Warping (DTW)~\cite{salvador2007toward}. The metric computes the minimum pairwise DTW distance between each trajectory in the source scenario (either generated or ground-truth) and its closest trajectory in the target scenario. To ensure the robustness and to prevent from inflated scores caused by an excessive number of generated trajectories (where at least one generated trajectory might cover each ground-truth trajectory), the metric averages the distances in both generated-to-ground-truth and ground-truth-to-generated directions. Additionally, the distance is normalized by the frame rate (\textit{fps}) to provide a value independent to temporal resolutions. The Minimum Pairwise Dynamic Time Warping \textit{DTW} metric is defined as:
\begin{equation}
\begin{gathered}
    d_\textit{DTW} = \frac{1}{2} \Big( \frac{d_{\mathcal{V}_\textit{Gen} \rightarrow \mathcal{V}_\textit{GT}}}{\textit{fps}} + \frac{d_{\mathcal{V}_\textit{GT} \rightarrow \mathcal{V}_\textit{Gen}}}{\textit{fps}} \Big), \smallskip\\
    where~~~~~ d_{\textit{Source}\rightarrow\textit{Target}} = ~~~~~~~~~~~~~~~~~~~~~~~~~~~~~~~~~~~~~~~~~~~~~~~~~~~\\
    ~~~~~~~~~~~~~~\frac{1}{N^\textit{Source}}\sum_{i=1}^{N^\textit{Source}} \min_{j\in[1, ..., N^\textit{Target}]} \!\!\!\textit{DTW} \big(\mathcal{T}_i^\textit{Source}, \mathcal{T}_j^\textit{Target} \big).
\end{gathered}
\end{equation}

\medskip
\vspace{1mm}\noindent\textbf{Diversity (\textit{Div.})}\quad
The diversity metric quantifies how unique the trajectories in the generated scenario are relative to the ground-truth scenario. It evaluates the diversity by calculating the number of trajectories from the source scenario that match to most similar trajectories in the target scenario. Similar to the \textit{DTW} metric, the \textit{Div.} metric averages the results from both generated-to-ground-truth and ground-truth-to-generated directions. The Diversity (\textit{Div.}) metric is defined as:
\begin{equation}
\begin{gathered}
    \textit{Div.} = \frac{1}{2}\big( \mathcal{J}_{\mathcal{V}_\textit{Gen} \rightarrow \mathcal{V}_\textit{GT}} + \mathcal{J}_{\mathcal{V}_\textit{GT} \rightarrow \mathcal{V}_\textit{Gen}} \big), \smallskip\\
    where~~~~~ \mathcal{J}_{\textit{Source}\rightarrow\textit{Target}} = ~~~~~~~~~~~~~~~~~~~~~~~~~~~~~~~~~~~~~~~~~~~~~~~~~~~\\
    ~~~~\frac{1}{N^\textit{Source}}\sum_{i=1}^{N^\textit{Source}} \mathbb{I}\big(i =\!\!\! \argmin_{j\in[1, ..., N^\textit{Target}]} \!\!\!\textit{DTW} \big(\mathcal{T}_i^\textit{Source}, \mathcal{T}_j^\textit{Target} \}  \big).
\end{gathered}
\end{equation}

\medskip
\vspace{1mm}\noindent\textbf{Collision Rate (\textit{Col.})}\quad
The percentage of test cases where the trajectories of different agents in the generated scenario run into collisions. We define collisions when the two agents are closer than 0.2 meters~\cite{liu2021snce}. The collision rate is defined as:
\begin{equation}
    \textit{Col.} = \frac{100}{T_{\mathcal{V}_\textit{Gen}}N} \sum_{t=0}^{T_{\mathcal{V}_\textit{Gen}}\!\!-1} \sum_{i=1}^{N} \mathbb{I} \Big( \exists j \neq i : \lVert \mathbf{c}_t^i - \mathbf{c}_t^j \rVert_2 < 0.2 \Big).
\end{equation}

\medskip
For evaluation, all datasets are resampled to match the 5 fps setting of the crowd behavior generation benchmark. During metric computation, we normalize the time intervals to minimize the effect of the frame rate. In specific, for the computation of \textit{Dens.}, \textit{Freq.}, \textit{Cov.}, and \textit{Pop.}, EMD is calculated after downsampling both target and generated scenarios to 1-second intervals. In the case of \textit{DTW}, the DTW distances are normalized by the frame rate. For \textit{Kinem.}, each component is normalized by the mean value of the corresponding ground-truth scenarios before EMD calculation.

\end{document}